# Visualization and clustering by 3D cellular automata: Application to unstructured data

Reda Mohamed HAMOU[1], Abdelmalek AMINE[2], Ahmed Chaouki LOKBANI[3] and Michel SIMONET[4]

[1,2,3] Taher Moulay University of Saïda, Algeria

[4] Joseph Fourier University, Grenoble, France

*Abstract*— **Given the limited performance of 2D cellular automata in terms of space when the number of documents increases and in terms of visualization clusters, our motivation was to experiment these cellular automata by increasing the size to view the impact of size on quality of results. The representation of textual data was carried out by a vector model whose components are derived from the overall balancing of the used corpus Term Frequency – Inverse Document Frequency (TF - IDF).The WorldNet thesaurus has been used to address the problem of the lemmatization of the words because the representation used in this study is that of the bags of words. Another independent method of the language was used to represent textual records is that of the n-grams. Several measures of similarity have been tested. To validate the classification we have used two measures of assessment based on the recall and precision (f-measure and entropy). The results are promising and confirm the idea to increase the dimension to the problem of the spatiality of the classes. The results obtained in terms of purity class (ie the minimum value of entropy) shows that the number of documents over longer believes the results are better for 3D cellular automata, which was not obvious to 2D the dimension. In terms of spatial navigation, cellular automata provide very good 3D performance visualization than 2D cellular automata.**

## I. Introduction

The exponential growth in the number of documents circulating the web has caused a rapid growth of activity on the Web and an explosion of data resulting from this activity. These data are at a very high percentage of text data that currently represent the support of the information used in web and communication, to do this, a tool of management, organization and analysis of these data is required. This new tool and algorithms had to be proposed to meet the growing needs in terms of extraction and interpretation of knowledge from these data. Data mining or specifically text mining that represents the search of textual data is thus growth in the development of techniques for extracting, manipulating and to formatting the information from the data, so in effect this is to automatically extract knowledge from a set of texts (corpus or database)using methods of discovery such as statistics, learning or data analysis. In our study, we are interested in a non-supervised learning problem in which a classification of textual documents is necessary. We have used a bio method inspired the cellular automata that we have already tested with the two dimension (2D) [1,2,3], but in the current study the experimentation is carried out with the three dimension (3D) because we wanted to solve the problem of the spatiality of the structure (do the non-supervised a large number of textual documents classification to large 2D cellular automaton to well represent the resulting clusters) and exploit the power of visualization of cellular automata 3D and its structure. Our work consists of three parts namely, a representation of the data portion, part classification and in the late part of visualization of classes.

### A. Problematic

The supervised classification problems are:
• Supervised classification requires a lot of human resources.
• When the number of documents increases, the initial classification is to review resulting in the creation of new classes and change of the criteria of choice.
• Little consistent classifications

Given the problems that the supervised classification pauses and according to [6] where about 80% of the documents are in text format. This huge volume of unstructured or structured semi gives rise to an act to find relevant information more difficult to achieve, generating a problem known as the problem of information overload [4]. The techniques and tools of knowledge discovery in text (KDT) [7] or simply text mining [6] are being developed to deal with this problem. One of these techniques is the clustering, a technique used to group similar documents of a given collection by assisting in the understanding of its content [8, 9]. One of its objectives is to similar documents in the same group and to place different documents in different groups. The assumption is that, thanks to a process of clustering, similar objects remain in the same group according to the attributes they have in common. This hypothesis is known as the hypothesis of cluster described in [10]. It is important to analyze the difference between clustering and classification processes. Both activities have different purposes and are different in essence: the process of grouping or clustering produces clusters without prior on the content of the document knowledge or which exist classes, while the classification (or categorization) is a process that begins with a predefined set of categories and tries to identify to which category a document belongs [9]. Thus, the clustering helps people to identify classes and build taxonomies. Some advantages of clustering are:

• Ability to consolidate large amounts of documents: humans cannot parse many aspects (characteristics) of documents.

• The impartiality in the process: human beings have a knowledge base (KB) which tends to bias the process of personal consolidation.

• Identification of the common features between documents (models such as words or concepts) : they can be used to understand why these documents in a same cluster are similar and include a collection of documents and its content.

• Organization of unstructured documents: clustering minimizes the overload of information (inability to analyze excessive amounts of information) in search of information, grouping similar information (sharing a set of documents), summarizing common characteristics, thus contributing to the user to view the results of a query or search, and help the user navigate through a set of documents automatically.

In this study we used and experienced unstructured data in this case the text data. Grouping or clustering of textual documents, in particular Web pages is one of the challenges of current research. This task is similar to the classification of documents structured in a database.

*B. State of the art*

A well designed clustering algorithm follows generally the four phases of design: the representation of the data, modelling, optimization, and validation [11]. The representation of the data phase predetermines what kind of cluster structures can be found in the data. On the basis of the representation of the data, the modelling phase defines the concept of clusters and criteria that separate structures desired groups of those non-desired or unfavourable ones. In this phase, a measure of quality which can be optimized or approached while searching for hidden in the data structures is produced. In General, traditional clustering algorithms can be classified into two categories: Hierarchical algorithms and Partitioning algorithms. There are two types of hierarchical algorithms: the hierarchical division algorithms and hierarchical algorithms of agglomeration. In a hierarchical division algorithm, the algorithm performs from the top to the bottom, that is to say, the algorithm starts with a large cluster containing all data in all of data points and continues to make splitting of clusters. In algorithm of hierarchical agglomeration, the algorithm proceeds from bottom to top, namely, the algorithm starts with clusters each containing a data point and continues to merge clusters. Unlike the hierarchical algorithms, partitioning algorithms create a level of non-overlapping partitioning of the data points. For large data sets, the hierarchical methods become impractical unless other techniques are integrated, because usually prioritised methods are of complexity O $(n^2)$ for space memory and O $(n^3)$ for time CPU [12, 13, 14], where n is the number of data in the dataset. Partitioning clustering algorithms attempt to break down all data directly in a series of disjointed groups. They try to optimize a certain test function which can focus on the local structure of the data. As a general rule, criteria of bases are to minimise some measure of dissimilarity in samples of each cluster, while maximizing the dissimilarity of the different clusters. The advantages of hierarchical algorithms are the disadvantages of partitioning algorithms and vice versa. Partitioning clustering algorithms are usually iterative in nature and converge Optima premises and include the k-means, k-médoïdes, Fuzzy C-Means, expectation - maximization (EM), k-harmonic means. What really interests us is a State of the art on the clustering using biomimetic methods since our method based on 3D cellular automata which is no other than a biomimetic method. The different types of algorithms which contributed in the field of classification data are: genetic algorithms, evolutionary programming, evolutionary, genetic programming strategies, the immune systems and intelligence in swarm in a comprehensive way. The idea of using genetic algorithms in the area of clustering has its forces to create a population of solutions candidates to an optimization problem, which is iteratively refined by alteration and selection of the right solutions for the next iteration. The candidate solutions are selected according to a fitness function that evaluates their quality. An advantage of these algorithms is their ability to face a local optimum by maintaining, recombination and comparison of multiple solutions simultaneously. In contrast, the heuristics of local research, such as the algorithm of simulated annealing [15], allows refining a single candidate solution and shows a weakness to deal with a local optimum. Determinism of local search, which is used in algorithms such as k-means, always converges to the nearest local optimum of the starting position of the research. Evolutionary or evolution strategies (ES) [16, 17] are optimization techniques based on ideas of adaptation and evolution. The evolution of strategies uses natural representations of dependent problems, and primarily mutation and selection of search operators. As algorithms evolutionary, the operators are applied in a loop. An iteration of the loop is called a generation. The sequence of generations is continued until a decision criterion is reached. The selection in evolution strategies is deterministic and is based only on rankings fitness not on values actual fitness. In the case of the clustering, Lee and Antonsson [18] used an evolution strategy (ES) to a set of data without any prior knowledge of the number of clusters. SLS proposed implements variable length individuals to simultaneously search for centroid and the number of clusters. Genetic programming (GP), is a methodology of evolutionary algorithm inspired by biological evolution to find computer programs that perform a task set by the user. It is a specialization of genetic algorithms, where each individual is a computer program. The main difference between the genetic programming and genetic algorithms is the representation of the solution. Genetic programming creates computer programs in computer languages Lisp or Scheme as a solution. Genetic algorithms create a string of numbers that represent the solution. The first results on the GP methodology was developed by Smith [20] and Cramer [21]. Koza, however remains the main initiator of the GP and paved the way for the application of genetic programming in complex optimization and various problems of research [19]. The work of Sarkar and al. [22] highlight an approach for classifying data dynamically using evolutionary programming (PE) where two fitness functions are optimized simultaneously. One gives the optimal cluster number, while the other leads to an identification of each cluster centroïd. This algorithm determines the optimal number of clusters and optimal cluster centers such that locally optimal solutions are avoided. Another advantage that this approach is that clustering is here independent of the initial choice of cluster centres in other words independent of initial conditions. However, the proposed method is applicable for the clustering of tasks where the clusters are not overlapping. For immune systems the Data: d1,...,dn represent antigens. These antigens are presented to the system repeatedly until a stop condition. The modelling data is digital, and therefore the antigen is a vector of dimension n. At each iteration, the antigen is presented to enable antibodies (similar, in this modelling, lymphocyte -B). An antibody is also represented by a vector of dimension n. The antibodies close enough to the antigen (in the sense of Euclidean distance) (ie: similar) will undergo clones with mutation (interaction antibody / antigen) to amplify and refine the system response. Also, these antibodies will undergo a selection (interaction antibody / antibodies): those who are too close to each other will be reduced in number. After these iterations, the system converges by placing antibodies (which act as sensors) wisely and number of suitable data. The main ideas used for the design of metaheuristics are operated upon the selections on lymphocytes, accompanied by the positive feedback allowing the multiplication and the memory system. Indeed, these attributes are critical to maintain the properties of self-organized system. With regard to swarm intelligence algorithms found a contribution in the area of clustering in ants and swarm particulate matter such as insects and birds. The behaviour of an Ant, bee, termite and the Wasp is often too simple, but their collective and social behaviour is of paramount importance. Approach to optimization by colonies of ants (Ant Colony Optimization) is a meta heuristics based on the population that can be used to find approximate solutions to difficult optimization problems. The main fields of applications of the COA are the problems of combinatorial optimization NP hard [23].

Since data classification problems and in particular the clustering is one of these problems then COA was one of the approaches to venture into this area. The particle swarm optimization (PSO) was introduced by Kennedy and Eberhart in 1995 [24]. They were inspired by research and simulation of behavior and the social psychology of insects. Research efforts have used to describe clustering as an optimization problem. This view gives us a chance to apply the PSO algorithm to evolve a set of cluster centroids candidates and thus determine a near optimal partitioning of the data at hand set. The clustering algorithm based on the PSO was introduced by Omran and al. [25]. [25, 26] Results show that the PSO-based method carries out the k-means, the FCM and a few other algorithms of classification on the State of the art. There are also other algorithms of clustering such as algorithms of clustering based on grid and density.

## II. THE REPRESENTATION OF DATA

The vector space of information search system, launched by Gerard Salton [27], [28], represents the documents as vectors in a vector space. The set includes a mxn Word document and a matrix document (A) , in which each column represents a document, and each A (i, j) entry represents the weighted frequency of the term i in document j. A major advantage of this representation is that the algebraic structure vector space can be exploited [29]. To achieve greater efficiency in the handling of data, it is often necessary to reduce the dimension especially when the data set is huge. The classic representation of a document text search for information is the representation in words bag (bag is a set where the repetitions are allowed), also known as the model vector (Vector Space), because a bag can be represented as a vector recording the number of occurrences of each word in the dictionary in the document at hand. In the vector model, a document is represented by a vertical vector indexed by all the elements of the dictionary (nth element of the vector is the ith term frequency in document TFi). A body is represented by a matrix D, whose columns are indexed by the documents and whose lines are indexed by terms, $D = (d_1,..., d_n)$. All of the terms of the collection have the same importance to determine the similarity (concept which we will later examine in detail) between the documents; we introduce the weight of the words. The weight of the term corresponds to the importance of the term in the corpus, and each element of the document vector is multiplied by the respective weight of the term. The most widely used weighting is called TF - IDF weighting and is used in our study [30]. IDF weight for term i of the dictionary is defined as follows: $Wi = \log (N / ni)$ Where, Ni is the number of documents in the corpus contains the word i.

Vector TF - DF of a document is a vector with the elements: $(Tf\text{-}Idf)_i = TFi \times \log (N / n_i)$. Bag of words representation excludes any form of grammatical analysis and any notion of distance between the words. Other representation, which has several advantages (essentially, this method processes the text regardless of the language

used), is based on the "n-grams (one "n-gram is a sequence of n consecutive characters")."

A document D is converted to a vector V in the following manner:

The idea is to encode each word of the bag by a scalar (number) called tf-idf to give a mathematical aspect to text documents. Where:
- $tf(i,j)$ is the term frequency: the frequency of term $t_i$ in document $d_j$
- $idf(i)$ is the inverse document frequency: the logarithm of the ratio between the number N of documents in the corpus and Ni the number of documents containing the term ti.

A document corpus di after vectorization is:
$d_i = (x_1, x_2, ... ...., x_m)$ where m is the number of word of $i^{th}$ bag of word and $x_j$ is the tf-idf.

### A. Dimension reduction

Several dimension reduction methods are described in the literature, but the methods of interest are those relating to the textual data is unstructured data used in our study. For a body of very large document, the treatment of the vector space representing such volume of data intends to apply a computation time and a very large memory space and may even prevent the use of classification algorithms. By definition, a text document is a set of words. Use these words for a task of classification for example could negatively influence on the accuracy of the classification. Indeed, and regardless of the language used several words are empty of meaning and therefore their use is unnecessary to emphasize the semantics of a text. In addition to these words of meaning, if a word appears in several documents of a corpus then it may not affect the membership of a document nor the document no-membership to a certain class, we then say that the power of the word discrimination is low. This dimension of vocabularies reduction techniques are born and are divided into two families: The selection of attributes («feature selection») and retrieving attributes ("feature extraction").

The principle of selection of attributes to reduce the dimension is very simple in its conceptual point of view and is to retain the attributes (or words) found to be relevant to the task of classification according to an evaluation function. The words that do not meet the assessment criteria are eliminated. These methods include the deletion of empty words, radical research (Stemming), frequency, information gain, the statistics of the χ2 and mutual information. The phenomenon of attribute extraction is to create a set of attributes that maximizes the classification process from the original attributes. We have techniques for extracting attributes, grouping of terms ("term clustering") and latent semantic indexing techniques. For our testing, we used the χ2 statistics and of gain of information because and from literature this two dimension reduction techniques give best results over the others.

*The information gain*

This measure reflects mathematically by:

$$G(t) = -\sum_{i=1}^{m} Pr(c_i) \log Pr(c_i) + Pr(t) \sum_{i=1}^{m} Pr(c_i|t) \log Pr(c_i|t) + Pr(\bar{t}) \sum_{i=1}^{m} Pr(c_i|\bar{t}) \log Pr(c_i|\bar{t})$$

Where
Pr (x) represents the proportion of documents having the characteristic x.
This means that a document is part of the category above
t means that a document has the term t
$\bar{t}$ means that a document does not have the term t
m is the number of class or classes

*The χ2 statistics*

The principle of this reduction technique is as follows:
Given the matrix Nij occurrences of i in the j text words.
- Calculate frequencies corresponding IFJ by fij = Nij / N
- Calculate (ij) contribution to the the χ2 statistic:

$$X_{i,j}^2 = \frac{\left(N_{i,j} - \frac{N_{i.} * N_{.j}}{N}\right)^2}{\frac{N_{i.} * N_{.j}}{N}} = N * \frac{(f_{ij} - f_{i.} * f_{.j})^2}{f_{i.} * f_{.j}}$$

- Sort the table of χ2 in descending order
- Finally, determine the list of the k first words for each text for standardization (heuristic choice)

Table 1 shows the reduction of dimension for 1000 documents by Chi2 method.

TABLE II. : REDUCTION OF DIMENSION BY CHI2 METHOD

| Number of Doc :1000 | | |
|---|---|---|
| N-Gram | Number of term before the reduction | Number of term after the reduction |
| 2 | 632 | 535 |
| 3 | 6164 | 2341 |
| 4 | 25264 | 2760 |
| 5 | 64164 | 2836 |

### B. The similarity matrix

The similarity coefficient indicates the strength of the relationship between two data points [31]. More two data points are similar over the similarity coefficient is large. Is x = (x1, x2,..., xd) and y = (y1, y2,..., yd) two points of data in d dimensions. Then, the coefficient of similarity between x and there will be a certain function of their attribute values, namely s(x,_y) = s(x1,_x2,...,_xd,_y1,_y2,...,_yd).

The similarity is generally symmetric, i.e. s (x, y) = s (y, x).

A proximity matrix is a matrix that contains the indices of pairs of proximity of a set of data. Usually, proximity matrices are symmetric. In what follows, a proximity index refers to similarity or dissimilarity index.

Given a set of data D = {x1, x2,..., xn}, each object is described by a vector of characteristics d-dimensional, the distance d matrix is defined as follows:

$$M_{dist}(D) = \begin{pmatrix} 0 & d_{12} & \cdots & d_{1n} \\ d_{21} & 0 & \cdots & d_{2n} \\ \vdots & \vdots & \ddots & \vdots \\ d_{n1} & d_{n2} & \cdots & 0 \end{pmatrix}$$

Where dij = d (xi, xj) that represents a digital distance between xi and xj data.

The similarity to D matrix is defined as

$$M_{sim}(D) = \begin{pmatrix} 1 & s_{12} & \cdots & s_{1n} \\ s_{21} & 1 & \cdots & s_{2n} \\ \vdots & \vdots & \ddots & \vdots \\ s_{n1} & s_{n2} & \cdots & 1 \end{pmatrix}$$

Where Sij = s (xi, xj) is the similarity between xj and xi data.

The matrix remote Mdist (D) and the similarity matrix Msim (D) the D data set are two examples of a proximity matrix. If the distance function and similarity function are symmetric then two proximity matrices are symmetric.

C. *The distances used*

The choice of the distances is important for applications, and the best choice is often achieved by a combination of experience, skills, knowledge, and opportunity. Here is a list of distances commonly used for digital data.

*Minkowsky distance of Order r*

$$d_{min}(\mathbf{x}, \mathbf{y}) = \left( \sum_{j=1}^{d} |x_j - y_j|^r \right)^{\frac{1}{r}}, \quad r \geq 1$$

When r = 1 is referred to as Manhattan distance or city block and for r = 2 is the Euclidean distance.

*Maximum distance*

The maximum distance is also called the "Sub" distance. It is defined as the maximum value of the distances of the attributes, two data points x and y in d dimensional space and is defined as follows:

$$d_{max}(\mathbf{x}, \mathbf{y}) = \max_{1 \leq k \leq d} |x_j - y_j|$$

*Average distance*

As noted by Legendre [32], the Euclidean distance has the following disadvantage: two points of data without values of attributes in common may have less than another pair of data points with the same attribute values. To overcome this drawback, the average distance was adopted amending the Euclidean distance [32]. the average distance is defined by:

$$d_{ave}(\mathbf{x}, \mathbf{y}) = \left( \frac{1}{d} \sum_{j=1}^{d} (x_j - y_j)^2 \right)^{\frac{1}{2}}$$

*Mahalanobis distance*

Mahalanobis distance can mitigate the distance of distortion caused by linear combinations of attributes. It is defined by:

$$d_{mah}(\mathbf{x}, \mathbf{y}) = \sqrt{(\mathbf{x} - \mathbf{y}) \Sigma^{-1} (\mathbf{x} - \mathbf{y})^T}$$

Where $\Sigma$ is the covariance matrix of the data set defined above.

*Tchebychev distance*

The Chebychev distance between two points is the maximum distance between the points in any single dimension. The distance between points X=($X_1$, $X_2$, etc.) and Y=($Y_1$, $Y_2$, etc.) is computed using the formula: Maxi |Xi - Yi| where Xi and Yi are the values of the ith variable at points X and Y, respectively.

III. MODELLING

Before modelling our clustering algorithm we detail our source of used data that represent the Reuters benchmark and a medical corpus.

A. *Reuters 21,758*

We used in our experiments the Reuters 21578 corpus, which represents a database of 21578 text documents of news information in English. The documents in the Reuters-21578 collection appeared on the Reuters newswire in 1987. The documents were assembled and indexed with categories by personnel from Reuters Ltd. In 1990, the documents were made available by Reuters and CGI for research purposes to the Information Retrieval Laboratory (W. Bruce Croft, Director) of the Computer and Information Science Department at the University of Massachusetts at Amherst. Formatting of the documents and production of associated data files was done in 1990 by David D. Lewis and Stephen Harding at the Information Retrieval Laboratory. The Reuters-21578 collection is distributed in 22 files. Each of the first 21 files (reut2-000.sgm through reut2-020.sgm) contains 1000 documents, while the last (reut2-021.sgm) contains 578 documents.

B. *medical Corpus*

This corpus consists of the OHSUMED collection which is extracted da set the MEDLINE database. It is composed of excerpts from medical journals dated from 1987 to 1991. In General, these texts contain a title and a summary, but some of them may have only a title. These texts contain, in addition to the manual annotations that correspond to manual categories called Medical Subject Headings MeSH.

C. *The cellular automaton 3D for Clustering*

We have already experienced the cellular automaton 2D [Hamou & al] and we have obtained satisfactory results, however we encountered a few problems of spatial representation of the classes (clusters) space. For this reason we have designed this problem by experimenting 3D cellular automata and we had reason because they provide a better representation of space and consume months of space to the 2D cellular automata.

Examples to make the clustering of 1000 documents and represent different classes there is a need of a cellular automaton size 2D 35 x 35, for the same number of documents is required of a cellular automaton 3D of size 11 x 11 x 11 (ie: a cube of dimension 11 cells) and its very expressive visual appearance.

From a formal point of view, a cellular automaton is defined by the quadruplet (U, V, E, F) where:

-U = (U1, U2,..., Ui,..., Un), finished a set of cells develops most often by a regular arrangement in a grid.

-V = V1, V2,..., Vi,..., Vn. all the neighbours of these cells, defined by a topological test (adjacency and / or connectivity) or metric that represents a distance between the centroid. Each set of neighbourhood Vi is a list of cells Vi = (Uj, Uk,..., Um) considered to be the neighbors of the i cell.

-E, all possible States of the cells as qualitative or quantitative, discrete or continuous variables.

-F, all the functions of local transition, deterministic or probabilistic, that allow evolving at each time step t, the State of cells based on the State of the cell and the States of the neighbouring, such as cells

$$E_i^{t+1} = f(E_i^t; E_{v_i}^t)$$

One usually distinguishes in this architecture, a structural component (U, V) and a dynamic component (E, F):

The cellular automata that we propose is a network of cells in a 3D space that belongs to the family (k, r) where k is the number of possible States of a cell to the cardinal of all States, and r is the environment of the cell either r is the radius of neighbourhood.

$$AC = \begin{matrix} Structure \\ (\{U,V\} \end{matrix}, \begin{matrix} Processus \\ \{E,F\}) \end{matrix}$$

U = {set of cells of a grid 3D}

This automaton has 4 possible States (k = 4) and the radius of neighbourhood is a single cell (r = 1).

V = {set of all the neighbours}

A cell of the automaton is dead, alive, isolated or contains data all States of the automaton are represented in the following E set:

E = {dead, alive, isolated, Active}

*Transition function [1,2,3]*

Almost the same 2D cellular automata transition functions are used for 3D cellular automata, the only difference is in the neighbourhood that contains a much larger number than that of the 2D cellular automata, namely:

The cellular automaton we proposed is a network of cells in a 3D space and belongs to the family (k, r) where k is the number of possible states of a cell i.e. the cardinal of all states and r is the environment of the cell i.e. r is the radius of the neighbourhood.

This automaton has 4 possible states (k = 4) and the radius of neighbourhood is a single cell (r = 1).

Thus a cell of the automaton is dead, alive, and alone or contains a data that all states of the automaton is (dead, alive, isolated, Active).

A dead cell will contain the value 0, a living cell will contain the value -1, an isolated cell will contain the value -2 and an active cell contain data (number of the document corpus).

We used these values and especially the value of the living cell (-1) to make the difference between a living cell containing the value 1 and a cell containing a data (number of the document) 1. Thus a cell will contain a value of {-2, -1, 0, 1, 2... N} where N is the number of the last document of the corpus used.

Rule 1:

If the cell $C_{i,j,k}$ dies  when then cell $C_{i,j,k}$ ← data
                    Neighbourhood $C_{i,j,k}$ becomes alive

Rule 2:

If cell $C_{i,j,k}$ Living
    Then Check Neighbourhood
        If neighbourhood contains at least one activate cell
            Then
                $C_{i,j,k}$ ← similar data
                Neighbourhood $C_{i,j,k}$ becomes alive
            Else
                Neighbourhood $C_{i,j,k}$ becomes isolated
        End
    End

Rule 3:

If a cell is isolated then unchanged (Isolated remains)

*The neighbourhood*

27 Blue cells (26 cells + the cell itself) are adjacent cells of the green cell in the vicinity of 3D Moore .(Figure 1).

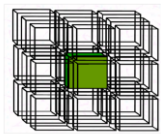 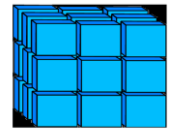

Figure 1.    3D Moore neighbourhood

7 Blue cells (6-cell + the cell itself) are adjacent cells of the green cell in the vicinity of 3D Von Neuman (Figure 2)..

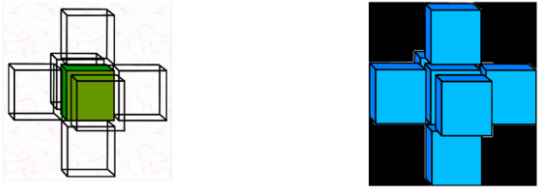

Figure 2. 3D Von Neumann neighbourhood

The neighbourhood used in our 3D cellular automaton is so that the spatial representation of the classes is quite airy namely 3D Moore neighbourhood:

## IV. EXPÉRIMENTATION

After testing our algorithm on the outcome of the Reuters 21578 corpus and medical corpus, we obtained the following results in terms of number of classes and purity of the clusters:

In terms of purity of the cluster, and error rate in classification we used two measures of assessment in this case the entropy and f-measure. Both measures are based on two concepts: recall and precision defined as follows:

$$precision(i,k) = \frac{N_{i,k}}{N_k}$$

$$recall(i,k) = \frac{N_{i,k}}{N_{c_i}}$$

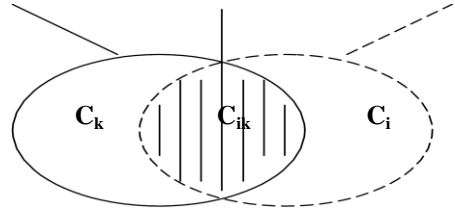

where N is the total number of documents, i is the number of classes (predefined), K is the number of clusters unsupervised classification, NCi is the number of documents of class i, Nk is the number of documents to cluster Ck, Nik is the number of documents of class i in cluster Ck. The entropy and f-measure are calculated on a partition P as follows:

$$E(p)=\sum_{k=1}^{K}\frac{N_k}{N}\times(-\sum precision(i,k)\times \log(precision(i,k)))$$

$$F(p)=\sum\frac{N_{C_i}}{N}\max_{k=1}^{K}\frac{(1+\beta)\times recall(i,k)\times precision(i,k)}{\beta\times recall(i,k)+precision(i,k)}$$

The partition P corresponds to the expected solution is one that maximizes the F-measure or minimizes the associated entropy. (In our study P is the partition that corresponds to the class of results of classification by the 2D Cellular Automata method for the number of documents associated).

### A. Results

TABLE II. CLASSIFICATION RESULTS OF 1000 DOCUMENTS BY 3D CELLULAR AUTOMATA WITH 3 DISTANCES SEPARATE

| | Cosine | | | | Euclidian | | | | Tchebychev | | | |
|---|---|---|---|---|---|---|---|---|---|---|---|---|
| | # Class | Time (ms) | E(p) % | F(P) % | # Class | Time (ms) | E(p) % | F(P) % | # Class | Time | E(p)% | F(p) % |
| 1 | 185 | 1845 | 5,50 | 70 | 162 | 1562 | 2,10 | 19 | 150 | 1452 | 2,10 | 19,70 |
| 2 | 114 | 1062 | 6,90 | 85 | 89 | 871 | 2,10 | 30 | 89 | 882 | 10,80 | 19,90 |
| 3 | 91 | 852 | 6,99 | 75 | 48 | 470 | 2,10 | 31 | 53 | 520 | 1,06 | 19,60 |
| 4 | 47 | 440 | 4,25 | 93,5 | 36 | 380 | 1,80 | 45 | 30 | 310 | 1,04 | 1,94 |
| 5 | 24 | 240 | 3,95 | 71,9 | 34 | 341 | 1,40 | 20 | 24 | 250 | 1,050 | 19,50 |
| 6 | 18 | 171 | 3,70 | 57,5 | 31 | 311 | 1,20 | 19,9 | 17 | 180 | 1,058 | 19,52 |
| 7 | 15 | 140 | 3,73 | 61,2 | 30 | 310 | 1,10 | 19,6 | 12 | 130 | 1,053 | 19,53 |
| 8 | 19 | 190 | 3,44 | 54,5 | 15 | 160 | 1,00 | 19,4 | 10 | 110 | 1,052 | 19,53 |
| 9 | 14 | 140 | 2,87 | 65,7 | 7 | 80 | 1,00 | 19,3 | 7 | 80 | 1,051 | 19,40 |
| 10 | 4 | 50 | 3,00 | 56,1 | 5 | 60 | 1,05 | 19,4 | 4 | 50 | 1,042 | 19,46 |

TABLE III. CLASSIFICATION RESULTS OF 1500 DOCUMENTS BY 3D CELLULAR AUTOMATA WITH 3 DISTANCES SEPARATE

| Seuil | Cosine | | | | Euclidian | | | | Tchebychev | | | |
|---|---|---|---|---|---|---|---|---|---|---|---|---|
| | # Class | Time (ms) | E(p) % | F(P) % | # Class | Time (ms) | E(p) % | F(P) % | # Class | Time | E(p)% | F(p) % |
| 1 | 175 | 3625 | 5,63 | 41,46 | 131 | 2854 | 1,53 | 15,78 | 136 | 2975 | 0,95 | 15,00 |
| 2 | 144 | 85 | 5,87 | 45,34 | 97 | 2103 | 1,50 | 15,53 | 110 | 2444 | 0,93 | 14,80 |
| 3 | 115 | 2404 | 4,97 | 25,18 | 62 | 1362 | 1,48 | 15,41 | 74 | 1642 | 0,46 | 14,83 |
| 4 | 89 | 1843 | 5,31 | 13,33 | 56 | 1231 | 1,49 | 32,44 | 63 | 1392 | 0,48 | 14,90 |
| 5 | 62 | 1292 | 4,69 | 39,00 | 49 | 1092 | 0,92 | 14,67 | 48 | 1092 | 0,46 | 14,78 |
| 6 | 39 | 811 | 2,57 | 76,51 | 36 | 801 | 0,92 | 14,68 | 37 | 841 | 0,48 | 15,09 |
| 7 | 27 | 581 | 2,12 | 91,53 | 28 | 631 | 0,47 | 14,90 | 29 | 661 | 0,47 | 15,03 |
| 8 | 18 | 401 | 1,98 | 48,76 | 23 | 531 | 0,47 | 14,86 | 16 | 370 | 0,47 | 14,95 |
| 9 | 10 | 241 | 3,12 | 78,55 | 15 | 350 | 0,46 | 14,80 | 11 | 260 | 0,47 | 14,91 |
| 10 | 7 | 170 | 1,80 | 42,17 | 10 | 241 | 0,48 | 15,03 | 5 | 131 | 0,47 | 104,87 |

TABLE IV. CLASSIFICATION RESULTS OF 226 DOCUMENTS BY 3D CELLULAR AUTOMATA WITH 3 DISTANCES SEPARATE

| Dist. Seuil | Cosine | | | | Euclidian | | | | Tchebychev | | | |
|---|---|---|---|---|---|---|---|---|---|---|---|---|
| | # Class | Time (ms) | E(p) % | F(P) % | # Class | Time (ms) | E(p) % | F(P) % | # Class | Time | E(p)% | F(p) % |
| 1 | 85 | 50 | 3,60 | 52 | 79 | 47 | 28,8 | 38,2 | 66 | 47 | 25,30 | 33 |
| 2 | 53 | 41 | 4,80 | 32 | 38 | 31 | 16,9 | 40,3 | 43 | 31 | 18,70 | 20,50 |
| 3 | 39 | 30 | 6,80 | 43,50 | 27 | 16 | 13 | 50,7 | 30 | 29 | 15 | 20,40 |
| 4 | 24 | 15 | 7,00 | 41,50 | 14 | 12 | 5 | 65,1 | 21 | 20 | 9 | 36,20 |
| 5 | 12 | 8 | 2,60 | 66,20 | 7 | 10 | 2,3 | 20,4 | 15 | 15 | 3 | 46,10 |
| 6 | 4 | 5 | 4,40 | 51,10 | 3 | 7 | 1 | 20,6 | 9 | 13 | 2,40 | 20,40 |

*Interpretation*

In table 2, we note that the cosine distance has a considerable advantage over other distances as part on grey background is the largest f-measure (this assessment criterion was chosen because this measure is based on the recall and precision). For classification with Euclidean distance, the best classification is represented on a yellow background in the table 2 and figure 4 and for the Chebychev distance results fail to meet our expectations and are those represented on blue in the table 2 and figure 5 because the f-measure is too far from the maximum value but a near entropy of the ideal value but the latter is based only on the precision. Regards figures 3,4,5,6,7 and 8 each color represents a class, and we note that the spatiality of the classes is done with ease and after navigation in the 3D cellular automata have identified the best possible plan. Based on Table 4 which shows the result of classification by cellular automata 3D medical corpus, we found that the number of classes for different distances was close to 13 since the medical corpus used was already categorized ( 13 classes) and there was a good compromise between the smallest and largest entropy f-measure. But the cosine distance performs best in terms of f-measure (66.20%) and entropy (2.60%). So we conclude that the cosine distance is recommended for experiments of cellular automata and text data but we still try other distances to give a final conclusion and just.

## B. VISUAL ASPECT OF 3D CELLULAR AUTOMATA

We have designed software composed of two parts. The first part is devoted to the classification and the second to the visualization. The software is fully configurable knowing that you can choose the colour of a cluster, the palette, the dimension of the scale and the functions of navigation (horizontal rotation, vertical rotation, zoom in and zoom out) to make a good visual classification. Each text document (data) is represented by a cube. A simple click of the mouse on the small cube allows us to view the corresponding gross document or its associated vector. Somehow our 3D cellular automatons like a magic cube. Software designed to make text clustering and visualization classes. This software consists of two parts, a data representation and part classification and visualization classes. The party representation receives as input a text corpus in which each document will be converted to numeric vector for each method chosen by the user in this case the n-grams and the conceptual approach (WordNet). The other part of the software, which is the largest in our study is devoted to the clustering of the vectors already found by calculating a similarity matrix represented by several distances quoted above and visualization clusters. The classification is performed by 3D cellular automata. These automata are visualized by a visualization system designed in JAVA 3D which offers very high performance navigation such as zooming in, zooming, rotation, horizontal and vertical rotation. Regarding the classification, two strategies constitution of clusters were developed in this case a constitution similar standard where documents are stored in neighboring cells (Von Neumann or Moore) and a linear formation where similar document agreed with next to the last document linearly. The figures below represent visualizations and classification results after navigation and an optimal Visual search for the different parameters of classification such as similarity threshold, the distance of similarity. Among other best classification results obtained in the tables above have been the work of optimization software also designed the study. Figures 3.4 and 5 respectively show the results of classification by cellular automata 3D cosine distance, Euclidean and Chebyshev. These figures show the best results in terms of validation tools (Entropy and F-measure) for different distances. After browsing the visualization platform based on cellular automata, we took the best plans. These jacks are made by the operations of rotation and zoom. Figures 7 and 8 show plans visualization of classes using the strategy of building linear clusters.

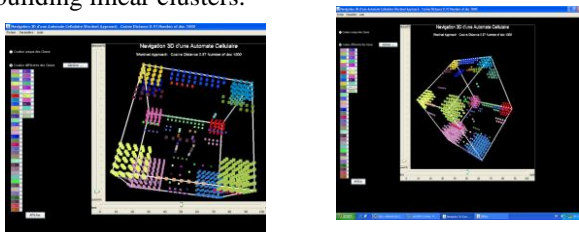

Figure 3 : Cosine Distance : 2 Views in the navigation of cellular automata

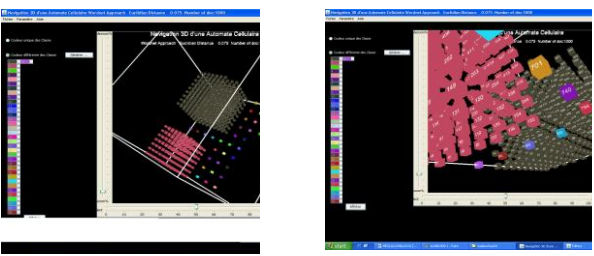

Figure 4 : Euclidian Distance: 2 Views in the navigation of cellular automata with Zoom

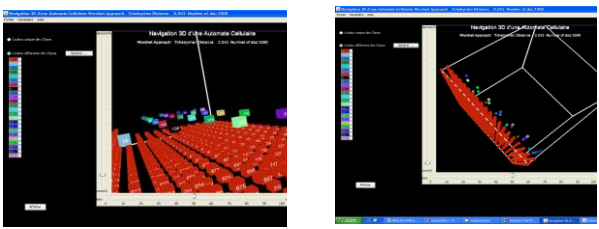

Figure 5 : Tchebychev Distance: 2 Views in the navigation of cellular automata with Zoom

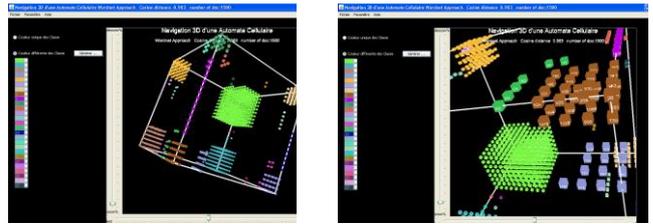

Figure 6 : Cosine Distance : 2 others views in the navigation of cellular automata

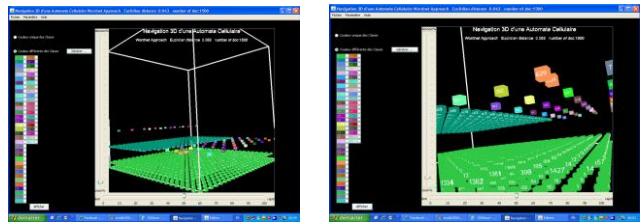

Figure 7 : Euclidian Distance: 2 Views in the navigation of cellular automata with Zoom right and an investment in the linear structure

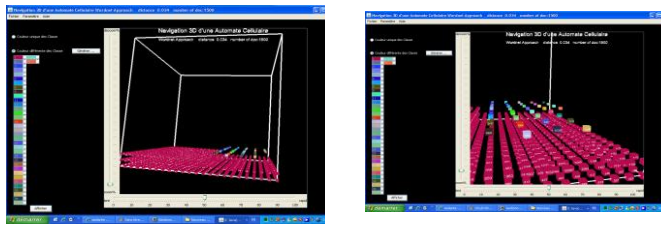

Figure 8 : Tchebychev Distance: 2 Views in the navigation of cellular automata with Zoom right and an investment in the linear structure

## V. CONCLUSION AND PERSPECTIVES

In this paper a 3D cellular automaton is proposed as a solution to the problem of unsupervised classification (clustering) of textual and spatial problems in classes in 2D cellular automata. The transition function used in our cellular automata evolved to form the group (cluster) similar to a certain threshold fields. The experimental results are positive and confirm the idea of increasing the dimension of the automat to navigate and make good reading classes in the visual design. The analysis and validation of classification results were based on evaluation criteria based on the notion of recall and precision are the f-measure and entropy. Given the results, our approach based on a biomimetic approach (3D cellular automaton) can help solve one of the problems of textual data mining and visualization. In perspective we will try to explore other methods biomimetic, because nature has not yet revealed all the secrets to solve combinatorial problems still existing in the field of data mining.